# Dark Spot Detection from SAR Images Based on Superpixel Deeper Graph Convolutional Network

Authors: Xiaojian Liu, Yansheng Li[*]

School of remote sensing and information engineering, Wuhan University, Wuhan 430079, China

Abstract:

Synthetic Aperture Radar (SAR) is the main instrument utilized for the detection of oil slicks on the ocean surface. In SAR images, some areas affected by ocean phenomena, such as rain cells, upwellings, and internal waves, or discharge from oil spills appear as dark spots on images. Dark spot detection is the first step in the detection of oil spills, which then become oil slick candidates. The accuracy of dark spot segmentation ultimately affects the accuracy of oil slick identification. Although some advanced deep learning methods that use pixels as processing units perform well in remote sensing image semantic segmentation, detecting some dark spots with weak boundaries from noisy SAR images remains a huge challenge. We propose a dark spot detection method based on superpixels deeper graph convolutional networks (SGDCN) in this paper, which takes the superpixels as the processing units and extracts features for each superpixel. The features calculated from superpixel regions are more robust than those from fixed pixel neighborhoods. To reduce the difficulty of learning tasks, we discard irrelevant features and obtain an optimal subset of features. After superpixel segmentation, the images are transformed into graphs with superpixels as nodes, which are fed into the deeper graph convolutional neural network for node classification. This graph neural network uses a differentiable aggregation function to aggregate the features of nodes and neighbors to form more advanced features. It is the first time using it for dark spot detection. To validate our method, we mark all dark spots on six SAR images covering the Baltic Sea and construct a dark spots detection dataset, which has been made publicly available (https://drive.google.com/drive/folders/12UavrntkDSPrItISQ8iGefXn2gIZHxJ6?usp=sharing). The experimental results demonstrate that our proposed SGDCN is robust and effective.

## 1 Introduction

Among all the various kinds of marine pollution, oil pollution ranks first in terms of occurrence frequency, extent of distribution, and degree of harm. The pollution of offshore waters caused by oil spills continues to occupy the attention of researchers in many countries. In particular, developed countries are investing a great deal of money in establishing oil spill monitoring systems to patrol, inspect, and manage offshore economic zones and territorial waters (Feng et al. 2014; Solberg 2012).

Synthetic Aperture Radar (SAR), which has the ability to penetrate clouds and fog and can work all day, is presently the most effective tool for oil pollution detection (Mera et al. 2012). Optical sensors can be used as a supplementary tool for oil pollution detection (Fingas and Brown 2017). As oil spills on the sea surface cause the attenuation of the Bragg waves and reduce the roughness of the sea surface, the oil film generally appears as dark spots on SAR images (Li and Li 2010). However, not all dark patches are caused by oil spills. The effects of atmospheric and oceanic phenomena, such as upwelling, ocean internal waves, rain cells, and low winds, also can appear as dark patches on SAR images, which are called "look-alikes", making it difficult to distinguish them from dark spots caused by oil spills (Shu et al. 2010).

---

[*] Corresponding author.
E-mail address: liuxiaojian2018@whu.edu.cn (Xiaojian Liu), yansheng.li@whu.edu.cn (Yansheng Li)

The traditional process of detecting oil pollution from SAR images is conducted in three phases: 1) dark spot segmentation, 2) feature extraction, and 3) dark spot classification (Shu et al. 2010). Dark spot segmentation is the first and most significant step of oil spill detection because its goal is to accurately segment all the dark spots in SAR images, including real oil spills and look-alikes. The accuracy of dark spot segmentation affects the quality of feature extraction and, ultimately, the accuracy of dark spot classification (Topouzelis 2008).

Although dark spot segmentation appears to be a simple binary classification problem, it is challenging due to the limitations of SAR sensors and the complexity of ocean conditions, as well as the characteristics of oil spills (Genovez et al. 2017; Konik and Bradtke 2016). There have been many studies pertaining to dark spot detection of oil spills using SAR, from simple to complex, which can be divided into three categories: 1) threshold-based approaches, 2) machine learning algorithms, and 3) deep learning algorithms. Threshold-based approaches are characterized by simplicity and fast calculation speed and usually require post-processing (Chehresa et al. 2016), such as the spatial density threshold method (Shu et al. 2010), the Otsu threshold method (Chehresa et al. 2016), and the multi-scale adaptive threshold segmentation algorithm (Solberg et al. 2007). Compared to the threshold-based approaches, machine learning algorithms are more intelligent in remote sensing image processing. A simple recurrent neural network proposed by Topouzelis et al. (2006) showed better performance in detecting dark formations. Taravat et al. (2014) developed a new dark spot detection approach from the combination of a Weibull multiplicative model and a pulse-coupled neural network, which proved to be fast, robust, and effective. Lang et al. (2017) designed three features suitable for dark spot segmentation: gray-scale features, geometric features, and texture features. The dark spot detection methods based on deep learning that emerged in recent years, such as Segnet (Guo et al. 2018), FCN (Cantorna et al. 2019; Guo et al. 2018), fully connected continuous conditional random field with stochastic cliques (Xu et al. 2016), and adversarial f-divergence learning (Yu et al. 2018), are capable of obtaining some better segmentation results, but still experience some errors and omissions.

Most of the above methods take pixels as the processing units. Although different strategies have been used to offset the influence of noise in SAR images, the results of dark spot segmentation are still unsatisfactory in some complex sea areas with high noise and weak boundaries. With the further development of superpixel segmentation algorithms, some studies have shown that they can be combined with deep learning methods to improve their performance (Uziel et al. 2019). Inspired by convolutional neural networks (CNNs), a large number of researchers began to develop more universal deep learning methods, called graph neural networks (GNNs), for processing non-euclidian data (Li et al. 2019; Liu et al. 2020). In this paper, we propose a dark spot segmentation method combining GNNs and superpixel segmentation. Compared with traditional pixel-based CNN algorithms and machine learning methods, our method has been shown to significantly improve segmentation performance. The details and results of our method are introduced later in this paper.

The contributions of this paper are as follows. 1) A dataset for dark spot detection in SAR images has been released, which can be helpful for future dark spot detection work. 2) For the first time, the deeper graph convolutional network is used for dark spot segmentation on SAR images. The images are transformed into graphs with superpixels as nodes, which serve as the input of the graph convolutional network. This method can smooth SAR image noise without changing the dark spot contours. 3) An optimal feature subset suitable for dark spot detection is proposed.

The remainder of this paper is structured as follows. Section 2 describes the study region and data used in this paper. Section 3 introduces our proposed dark spot detection method in detail, including images to graphs transformation, feature extraction, feature selection, and a GNN. Section 4 and Section 5 presents and discusses the research results, respectively. Finally, Section 6 presents our conclusions and future directions of work.

## 2 Data and study region

A collection of six Advanced Synthetic Aperture Radar (ASAR) products from the Envisat satellite was used to prove the effectiveness of the method we propose in this paper. Envisat ASAR was developed by the European Space Agency (ESA) and operated in the C band in a wide variety of modes. Table 1 shows the main characteristics of these modes. The wide swath mode (WSM) in VV polarisation we utilized in our method is a unique instrument for detecting oil slicks on the ocean surface because it offers a very good combination of wide coverage and radiometric quality (ESA 2007). Inevitably, there also may be some brightness effects between subswaths of the image acquired in wide swath mode, which may hinder the detection of oil slicks. Najoui et al. (2018) applies a semi-linear model to solve this problem. The images we used cover most of the Baltic Sea, which is an important waterway in Northern Europe. The marine environment and coastal ecology of the Baltic Sea are constantly threatened by oil discharges from ships (Konik and Bradtke 2016). The Baltic Sea images contain dark spots of various shapes and sizes, which are manually marked to form a dataset for dark spot detection. We make this dataset public.

Table 1. Main characteristics of the different modes.

| Mode | Id | Polarisation | Incidence | Resolution | Swath |
| --- | --- | --- | --- | --- | --- |
| Alternating polarisation | AP | HH/VV,HH/HV,VV/VH | 15-45° | 30-150m | 58-110km |
| Image | IM | HH,VV | 15-45° | 30-150m | 58-110km |
| Wave | WV | HH,VV | 15-45° | 400m | 5km×5km |
| Suivi global | GM | HH,VV | 15-45° | 1000m | 405km |
| Wide Swath | WS | HH,VV | 15-45° | 150m | 405km |

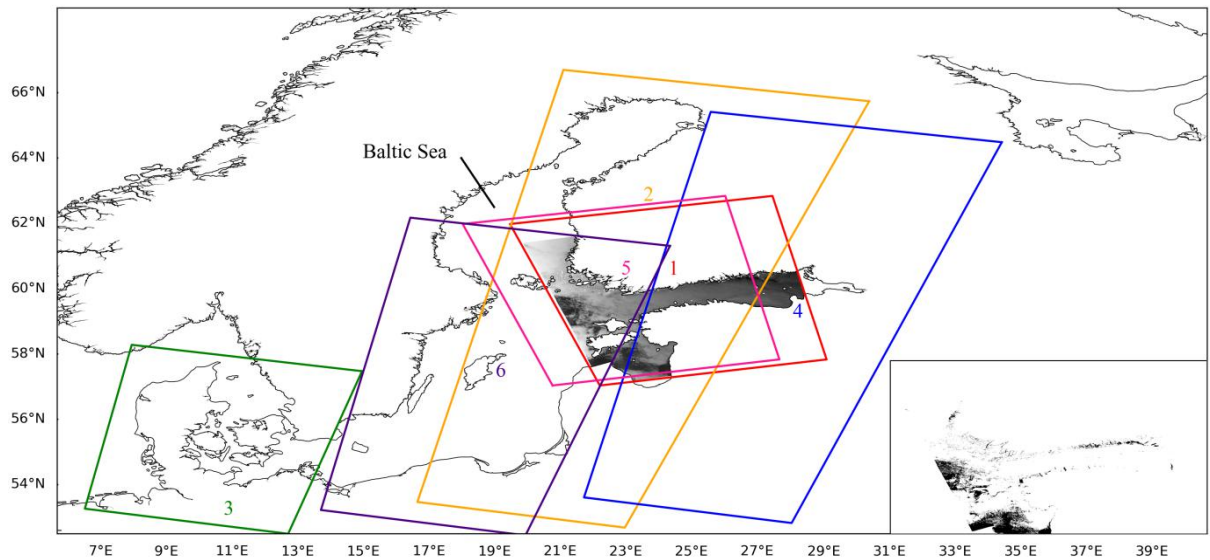

Figure 1. Overview of the research area. Numbers 1 through 6 show the coverage of the images used; and the artificially marked dark spots in image 1 are shown in the lower right corner.

## 3 Methods

The dark spot detection method we propose is comprised of three steps: 1) conversion of the image to graph structures, 2) feature extraction and selection, and 3) graph node classification. The details of each step are explained in the following sections.

3.1 Conversion of the image to graph structures

Before conversion, SAR images need to be preprocessed. It includes radiometric calibration, reprojection, speckle-filter, and masking out the land. We carried out preprocessing by using the sentinel application platform (SNAP), which is a common architecture for ESA Toolboxes (Misra and Balaji 2017). A 3×3 Lee filter was chosen for speckle filtering. The Lee filter has proved to be a very successful filter in the image processing of oil spill detection and has been applied many times (Genovez et al. 2017). After preprocessing, SAR images need to be segmented by superpixels. The use of superpixels has become a very popular technique in the applications of image processing (Giraud et al. 2018). Some researchers have indicated that the features calculated from superpixel regions were more robust than those from fixed pixel neighborhoods (Habart et al. 2017). Superpixel segmentation tries to reduce the noise of processing at the pixel level. By grouping the pixels into homogeneous regions, the number of elements to be processed is also reduced, thus drastically reducing the computational burden (Giraud et al. 2018). Konik and Bradtke (2016) showed that smooth SAR images or gradient images with obvious boundaries can significantly increase the accuracy of determining the outlines of dark spots. The Bayesian adaptive superpixel segmentation method (BASS) we use in this paper, which is the state-of-the-art method at present, supports massive parallelization and can be implemented in GPU (Uziel et al. 2019). After superpixel segmentation, the shapes, sizes, and number of adjacent superpixels are changed. In order to process them efficiently, adjacent superpixels are connected and images are transformed into graph structures with superpixels as nodes. Due to limited memory, the object of superpixel segmentation is an image cropped into small pieces. After transformation, we will get a large number of sub-graphic structures, which inevitably lose some edges. However, in practical applications, its impact on the accuracy of dark spot segmentation is very weak and can be ignored.

3.2 Feature extraction and feature selection

In previous studies, feature extraction was used mainly to extract the features of dark spots to distinguish between oil slicks and look-alikes (Genovez et al. 2017). Various features have been proposed for dark spot classification: Solberg et al. (2004) extracted 12 features; Vyas et al. (2015) used 25 features; Chehresa et al. (2016) extracted 74 features, and Mera et al. (2017) extracted 52 features. While the number of features was different in each of these studies, all of them can be divided into three categories: geometrical features, physical features, and textural features (Mera et al. 2017). In order to improve the performance of dark spot detection, in this paper, we propose extracting the features on superpixels, specifically 52 features, as proposed by Mera et al. (2017). However, the features related to the wind for dark spot classification were eliminated as well as features that were difficult to calculate when some of the superpixels only contained one pixel. Ultimately, we retained and linearly normalized 48 features before using them (Table 2). To reduce the difficulty of learning tasks, feature selection is required after feature extraction. Feature selection is a very important data processing procedure to alleviate the curse of dimensionality. In this paper, we chose to utilize support vector machines recursive feature elimination (SVM-RFE) for feature selection of superpixels (Mera et al. 2017). SVM-RFE is an embedded feature selection method. It works by iteratively training an SVM classifier, ranking the scores of each feature according to the weights of the SVM, removing the feature with the lowest score, and finally selecting the features required.

Table 2. The features used in our method.

| No | Feature | Code | Category | No | Feature | Code | Category |
| --- | --- | --- | --- | --- | --- | --- | --- |
| 1 | Area | A | Geometrical | 25 | Var_area_superpixel | Vas | Textura |
| 2 | Perimeter | P | Geometrical | 26 | Mean Haralick | H | Textura |
| 3 | Perimeter to area ratio | P/A | Geometrical | 27 | Object mean | Om | Physical |
| 4 | Area to perimeter ratio | A/P | Geometrical | 28 | Object standard deviation | Osd | Physical |
| 5 | Elongation | E | Geometrical | 29 | Background mean | Bm | Physical |
| 6 | Major axis to perimeter ratio | Maxx/P | Geometrical | 30 | Background standard deviation | Bsd | Physical |
| 7 | Complexity1 | Cp1 | Geometrical | 31 | Mean of the contrast ratio | Crm | Physical |
| 8 | Complexity2 | Cp2 | Geometrical | 32 | Standard deviation of the contrast ratio | Crsd | Physical |
| 9 | Circularity | C | Geometrical | 33 | Object power to mean | Opm | Physical |
| 10 | Spreading | S | Geometrical | 34 | Background power to mean | Bpm | Physical |
| 11 | Superpixel width | Sw | Geometrical | 35 | Ratio of the power to mean ratios | Opm/Bpm | Physical |

| 12 | Curvature | Cu | Geometrical | 36 | Max contrast | Cmax | Physical |
| 13 | Hu moments | Hu | Geometrical | 37 | Mean contrast | Cm | Physical |
| 14 | Fluser and Suk moments | Fs | Geometrical | 38 | RISDI | RISDI | Physical |
| 15 | Thickness | T | Geometrical | 39 | RISDO | RISDO | Physical |
| 16 | Shape connectivity | Shc | Geometrical | 40 | IOR | IOR | Physical |
| 17 | Form factor | Ff | Geometrical | 41 | Gradient mean | Gm | Physical |
| 18 | Length to width ratio | L/W | Geometrical | 42 | Gradient standard deviation | Gsd | Physical |
| 19 | Shape index | Si | Geometrical | 43 | Max. gradient | Gmax | Physical |
| 20 | Narrowness | N | Geometrical | 44 | Object border gradient | Obg | Physical |
| 21 | Rectangular saturation | Rs | Geometrical | 45 | Surrounding Power-to-mean ratio | Spm | Physical |
| 22 | Marking ratio | Mr | Geometrical | 46 | RIIA | RIIA | Physical |
| 23 | Solidity | Sd | Geometrical | 47 | Elliptic Fourier Descriptors | EFD | Geometrical |
| 24 | Mean of the interior angles based on bounding polygons | IABPm | Geometrical | 48 | Standard deviation of the interior angles based on bounding polygons | IABPsd | Geometrical |

3.3 Deep learning on Graphs

A graph convolutional network (GCN), which is a GNN variant, is a promising deep learning method that has been developed rapidly in the past few years. GCNs exploit message passing or equivalently certain neighborhood aggregation functions to extract high-level features from a node as well as its neighborhoods (Rong et al. 2020). GCNs are improving optimal results for a variety of tasks on graphs, such as node classification (Kipf and Welling 2017; Li et al. 2020a), linking property prediction (Zhang and Chen 2018), and graph property prediction (Lee et al. 2019).

A graph $G$ is usually defined as a tuple of two sets $G=(V,E)$. $V=\{v_1,v_2,...v_i,v_{i+1},...v_N\}$ and $E \subseteq V \times V$ are the sets of vertices and edges, respectively. $v_i$ represents the $i$-th node in the graph. If G is an undirected graph, the edge $e_{i,j}=(v_i,v_j) \in E$ indicates that the node $v_i$ is connected to $v_j$, otherwise, it means from node $v_i$ to $v_j$. $he_{vu}^{(l)}$ denotes edge features of node $v$ to $u$ in layer $(l)$. $h_v^{(l)} \in R^F$ is node features of node v in layer (l). Message passing by graph convolutional networks can be described as formula (1)-(3),

$$m_{vu}^{(l)} = \rho^{(l)}(h_v^{(l)}, h_u^{(l)}, h_{e_{vu}}^{(l)}), u \in N(v) \tag{1}$$

$$m_v^{(l)} = \zeta^{(l)}\left(\left\{m_{vu}^{(l)} \mid u \in N(v)\right\}\right) \tag{2}$$

$$h_v^{(l+1)} = \phi^{(l)}(h_v^{(l)}, m_v^{(l)}) \tag{3}$$

Where $\zeta^{(l)}$ is the message aggregation functions, which are differentiable, permutation invariant function, such as sum, mean, or max, $N(v)$ represents the set of neighbor nodes of $v$, and $\phi^{(l)}$ and $\rho^{(l)}$ denote differentiable functions, such as multi-layer perceptrons (MLPs).

DeeperGCN (Li et al. 2020a), an effective GCN, was chosen in this paper. Its message construction function $\rho^{(l)}$ and differentiable function $\phi^{(l)}$ are as follows:

$$m_{vu}^{(l)} = \rho^{(l)}(h_v^{(l)}, h_u^{(l)}, h_{e_{vu}}^{(l)}) = \mathrm{ReLU}(h_u^{(l)} + 1(h_{e_{vu}}^{(l)}) \cdot h_{e_{vu}}^{(l)}) + \varepsilon, u \in N(v) \tag{4}$$

$$h_v^{(l+1)} = \phi^{(l)}(h_v^{(l)}, m_v^{(l)}) = \mathrm{MLP}(h_v^{(l)} + s \cdot \left\| h_v^{(l)} \right\|_2 \cdot m_v^{(l)} / \left\| m_v^{(l)} \right\|_2) \tag{5}$$

Where ReLU(·) is the rectified linear unit and 1(·) is the indicator function, which is 1 when the edge feature exists or otherwise 0. $\varepsilon$ is the small constant and its value is $10^{-7}$, MLP(·) is a multi-layer perceptron, and $s$ is a learnable scaling factor. As shown in Equations (6) and (7), DeeperGCN uses several differentiable generalized message aggregation functions $\zeta^{(l)}$, which unify the different message aggregation operations, such as mean, max, and min. The optimal aggregation function can be automatically selected in each layer of DeeperGCN through training to aggregate the features of the superpixels and their neighbor nodes.

$$SoftMax\_Agg_\beta(\cdot) = \sum_{u \in N(v)} (exp(\beta m_{vu}) / \sum_{i \in N(v)} exp(\beta m_{vu})) \cdot m_{vu} \tag{6}$$

$$PowerMean\_Agg_p(\cdot) = \left(1/|N(v)| \cdot \sum_{u \in N(v)} m_{vu}^p\right)^{1/p}, P \neq 0 \tag{7}$$

Where $\beta$ and $p$ are the learnable variables. Formally, $\lim_{\beta \to 0} SoftMax\_Agg_\beta(\cdot) = Mean(\cdot)$, $\lim_{\beta \to \infty} SoftMax\_Agg_\beta(\cdot) = Max(\cdot)$, $\lim_{\beta \to -\infty} SoftMax\_Agg_\beta(\cdot) = Min(\cdot)$, $PowerMean\_Agg_{p=1}=Mean(\cdot)$, $\lim_{p \to \infty} PowerMean\_Agg_p(\cdot) = Max(\cdot)$, $\lim_{p \to -\infty} PowerMean\_Agg_p(\cdot) = Min(\cdot)$. Both of the above two functions can be generalized to the Sum aggregator by multiplying $y$-th power of the degree of vertexs, where $y$ is a learnable variable. In order to help to train the DeeperGCN architecture and improve performance, DeeperGCN uses a pre-activation variant of residual connection (Res+), which follows this ordering: Normalization→ ReLU→ GraphConv→ Addition (Li et al. 2020a).

# 4 Results

In this section, we compare the performance of our method on several dark spots (including oil slicks and look-alikes) with the following segmentation methods: 1) PROP (Lang et al. 2017), 2) Otsu+post-processing (Chehresa et al. 2016), and 3) Adversarial f -Divergence Learning (Yu et al. 2018) as well as two classic CNN methods: 1) Unet (Ronneberger et al. 2015), 2) Segnet(Guo et al. 2018)).

4.1 Experimental Setup

The pre-processed images were cropped into 5,030 samples with a size of 256*256 pixels. According to the ratio of 6:2:2, 2,898 samples were randomly selected for training, 1,022 samples for verification, and the remaining 1,019 samples for testing. The pixel ratio of dark spots to the background in the training sample was approximately 1:9.

In the superpixel segmentation, in order to reduce the computational burden, the object of BASS processing was the cropped images. The number of initial superpixels in BASS was set to be 3,000 to ensure that dark spots in the small areas could be divided into superpixel patches. The maximum number of iterations was set to 250, and the remaining parameter settings were the same as those in the previous study (Uziel et al. 2019). After feature extraction, machine learning library (scikit-learn) in Python was used to implement the SVM-REF feature selection algorithm with the default parameters.

In the graph node classification, we implemented our DeeperGCN model based on PyTorch Geometric, and used the Adam optimizer with an initial learning rate of 0.001. The hidden channel size was 128, the batch size was 16, and the number of GCN layers was 28. The dropout was 0.2 for MLP. Both $t$ and $s$ were initialized to 1.0, while $y$ was initialized to 0.0. The message aggregation function was $|N(v)|^y \cdot \text{SoftMax\_Agg}_\beta(\cdot)$.

Four indicators were used for quantitative evaluation: the detection probability ($P_d$), false alarm probability ($P_f$), overall accuracy ($P_{acc}$) (Lang et al. 2017), and the missing ratio of oil spill ($P_m$), which are defined as

$$P_d = TP/(TP+FN) \times 100\% \tag{8}$$

$$p_f = FP/(TP+FP) \times 100\% \tag{9}$$

$$p_{acc} = (TP+TN)/(TP+TN+FP+FN) \times 100\% \tag{10}$$

$$p_m = (MO/AO) \times 100\% \tag{11}$$

where TP (true positive) and TN (true negative) denote the number of pixels with correctly predicted dark spots and seawater, respectively; FN (false negative) and FP (false positive) refer to the number of pixels incorrectly predicted as seawater and dark spots, respectively; MO and AO are the number of pixels in missing oil spill areas and in all oil spill areas, respectively.

4.2 Effect of feature selection

After feature extraction, each superpixel received one eigenvector including 137 eigenvalues linked to the feature space presented in Section 3.2. In order to obtain a powerful subset of features for dark spot detection, we used the SVM-RFE algorithm for feature selection and ranked the 137 eigenvalues. The top 30 eigenvalues were selected as the optimal feature subset for dark spot detection because the F1 score of the SVM classifier was stabilized while using the top 30 eigenvalues (Figure 2). The codes and categories of the top 30 eigenvalues are shown in Table 3, which includes thirteen physical features, sixteen geometrical features, and one textural feature. Seven physical features and three geometric features were among the top 10 features, and the top five features were all physical features.

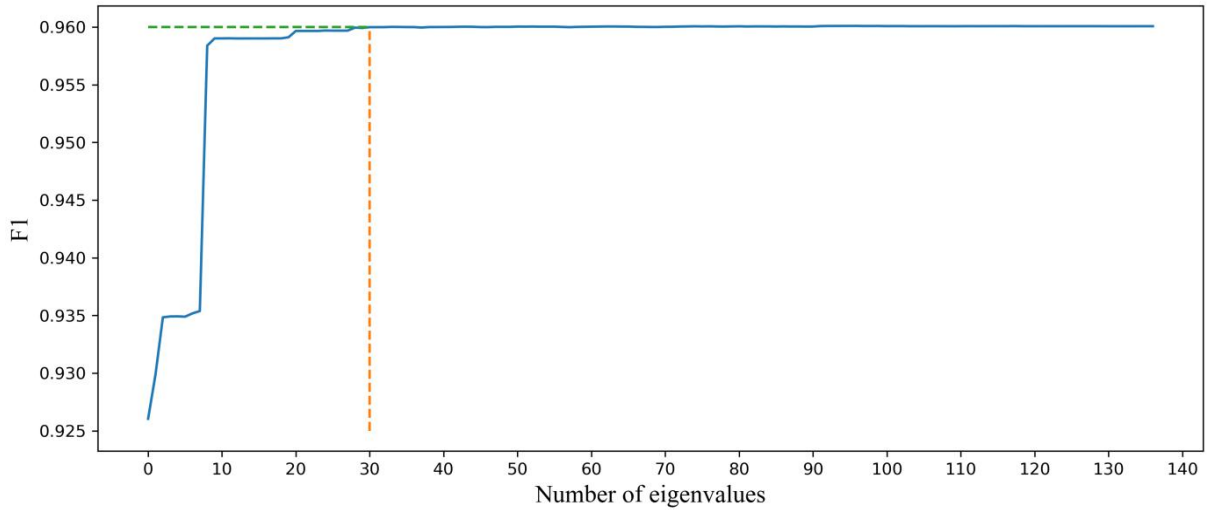

Figure 2. F1 scores corresponding to different eigenvalues.

Table 3. A powerful subset of features.

| Rank | Code | Category | Rank | Code | Category | Rank | Code | Category |
|---|---|---|---|---|---|---|---|---|
| 1 | RIIA | Physical | 11 | Fs4 | Geometrical | 21 | Bsd | Physical |
| 2 | Cm | Physical | 12 | Cp1 | Geometrical | 22 | IOR | Physical |
| 3 | Obg | Physical | 13 | Vas | Textural | 23 | Bpm | Physical |
| 4 | Gm | Physical | 14 | A/P | Geometrical | 24 | Bm | Physical |
| 5 | RISDO | Physical | 15 | Fs3 | Geometrical | 25 | Cp2 | Geometrical |
| 6 | A | Geometrical | 16 | RISDI | Physical | 26 | L/W | Geometrical |
| 7 | P | Geometrical | 17 | Spm | Physical | 27 | E | Geometrical |
| 8 | C | Geometrical | 18 | Rs | Geometrical | 28 | Si | Geometrical |
| 9 | Om | Physical | 19 | Sd | Geometrical | 29 | P/A | Geometrical |
| 10 | Osd | Physical | 20 | Mr | Geometrical | 30 | T | Geometrical |

In order to verify the validity of the selected feature subset, we used the selected feature subset (the top 30 eigenvalues) and all the features trained by the DeeperGCN classifier, respectively, and used the test dataset to verify the accuracy of the two training models. Due to the small number of oil spills in the test dataset, the missing ratio of oil spills of two trained models was calculated on another dataset containing 114 oil patches. The results of the training model using the selected feature subset were better than the results using all the features (Table 4), which means that the number of eigenvalues used for training the GCN was not as many as possibly could be used. The physical features played a major role in dark spot segmentation, followed by the geometric features, and finally the textural features. The feature subset we selected not only reduced considerably the number of features to be calculated and accelerated the feature extraction process, but also improved the accuracy of the model. Our proposed superpixel-based DeeperGCN model is abbreviated as SDGCN from here forward.

Table 4. Comparison of the top 30 eigenvalues with 137 eigenvalues.

| Model | $P_d$(100%) | $P_f$(100%) | $P_{acc}$(100%) | $P_m$(100%) |
|---|---|---|---|---|
| SDGCN with the top 30 eigenvalues | **96.98** | **5.68** | **99.16** | **7.18** |
| SDGCN with all eigenvalues | 95.74 | 6.68 | 98.95 | 8.73 |

4.3 Comparison with different segmentation methods

In this section, our proposed method SDGCN is compared with other classic pixel-based segmentation methods from a quantitative and qualitative perspective, including threshold methods, machine learning methods, and deep learning methods. Here, SDGCN adopts the top 30 eigenvalues.

Table 5. Comparison of the accuracy of different dark spot segmentation methods.

| Method | $P_d$(100%) | $P_f$(100%) | $P_{acc}$(100%) | $P_m$(100%) |
|---|---|---|---|---|
| Otsu+post-processing | 71.74 | 12.78 | 95.43 | 26.35 |
| PROP | 90.36 | 52.71 | 94.39 | 10.36 |
| SegNet | 83.00 | 8.88 | 97.35 | 13.03 |
| UNet | 83.20 | 6.69 | 97.52 | 9.68 |
| Adversarial f-Divergence Learning | 87.06 | 21.62 | 96.77 | 7.31 |
| SDGCN | **96.98** | **5.68** | **99.16** | **7.18** |

From the quantitative evaluation results of the various models shown in Table 5, we observed that the performance of SDGCN was significantly better than the other models, which demonstrated the superiority of the segmentation scheme we proposed. Converting the image to graphs with superpixels as nodes does improve the accuracy of dark spot detection. Nine images (256*256 pixels) with different brightness levels and a variety of dark spots are shown in Figure 3, and the results of each model were as follows. Otsu+post-processing achieved effective segmentation for dark spots with obvious boundaries; but when the boundary was not obvious, its segmentation results were not ideal. The PROP algorithm almost did not complete the segmentation of dark spots on the image with strong brightness because it is not able to consider the impact of the incident angle on the image brightness (Topouzelis et al. 2016). The segmentation results of Segnet and Unet were relatively close, but their segmentation performance was poor on the image with similar intensity for dark spots and backgrounds and also was affected by SAR image noise. Adversarial f-Divergence Learning performed worse than Segnet and Unet on darker images, but it did segment the dark spots accurately on brighter images. Compared with the other models, the results of SDGCN were the most similar to the truth labels, and the dark spots with blurred edges were accurately segmented.

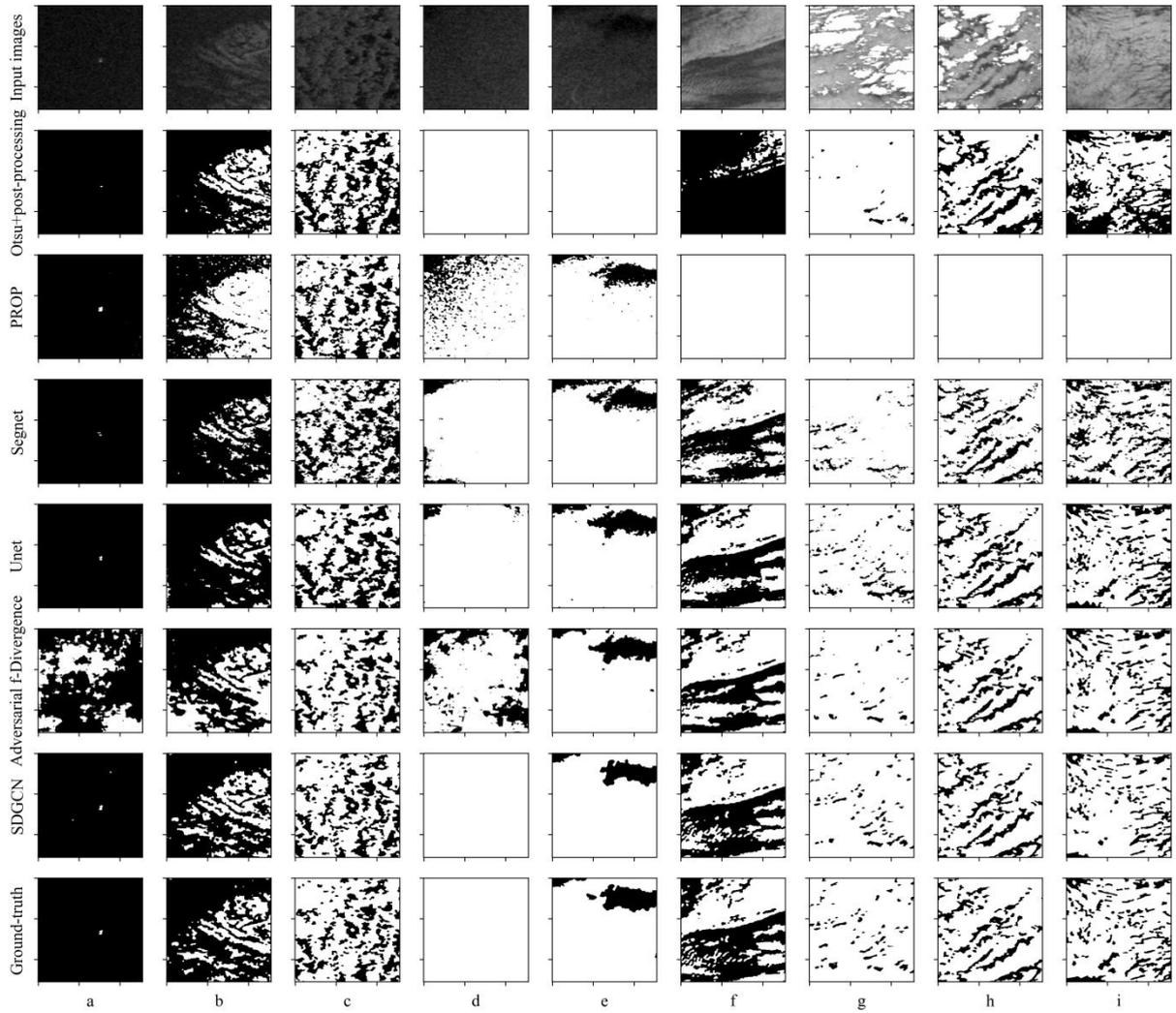

Figure 3. Several examples of qualitative comparison results; a ~ i are the different images used and their segmentation results.

## 5 discussions

In this section, we analyze the practicality of SGDCN method on another dataset containing 27 SAR images and illustrate the limitations as well as future work. In this dataset, each image contained at least one oil patch in locations published by the Baltic Marine Environment Protection Commission (Helsinki Commission - HELCOM). Because some of the oil patches were broken into tiny plaques by the waves, it was difficult to count them accurately. The length of time they appeared at sea depended on the amount of oil leaked. Other factors that affected the nature of oil patches and their characteristics in the images were different sea temperatures, salinity, ocean current speed, ocean waves, and atmospheric conditions. About 4,500 images with 256*256 pixels were used to retrain the SDGCN model, and the trained model segmented the dark spots on 27 SAR images.

Figure 4 shows the unsegmented oil patches in all the images. It can be seen that only the oil patches with weak borders and small areas shattered by waves were missed. One of the reasons is that small oil leakages can be greatly affected by advection, diffusion, evaporation, and emulsification (Berry et al. 2012). As a result, the nature of the oil changes, and they were not very different from the ocean background in the image, making it very difficult to detect them successfully. Another reason may have been that the number of training samples was small and the training sample labels were not very accurate, which may have led to the overfitting phenomenon of SDGCN during the training process.

Some of these detected oil patches of various shapes and sizes are shown in Figure 5. According to our statistics, a total of 103 pieces were detected, including small oil patches broken by the waves. Garcia-Pineda et al. (2009) determined that the optimum wind speed range to study the surface of oil slicks in SAR images was 3.5~7.0 m/s. We applied the CMOD5 geophysical model function to derive the wind speeds of the sea surface for oil spill analysis (Hersbach et al. 2005). We found that the wind speed on the sea surface of all the oil patches segmented ranged from 1 to 8 m/s. The smallest area of the detected oil patches was 0.2 km$^2$, and the largest was 245 km$^2$. In Figure 5, oil patches a, j, and k appear as dots with small areas, and oil patch m appears as a lump shape. In addition, other oil patches appear as long strips, almost all of which were formed by illegal discharge from ships. It can be seen that the segmentation results of SDGCN model basically look the same as the oil patches on the input image, indicating very good performance of oil patch segmentation.

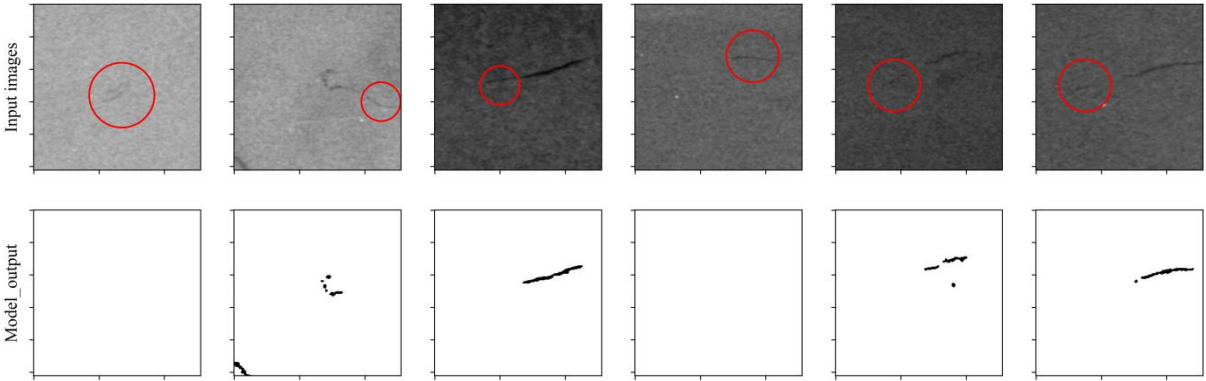

Figure 4. Oil patches that were missed; the red circles indicate the oil patches that were not successfully segmented.

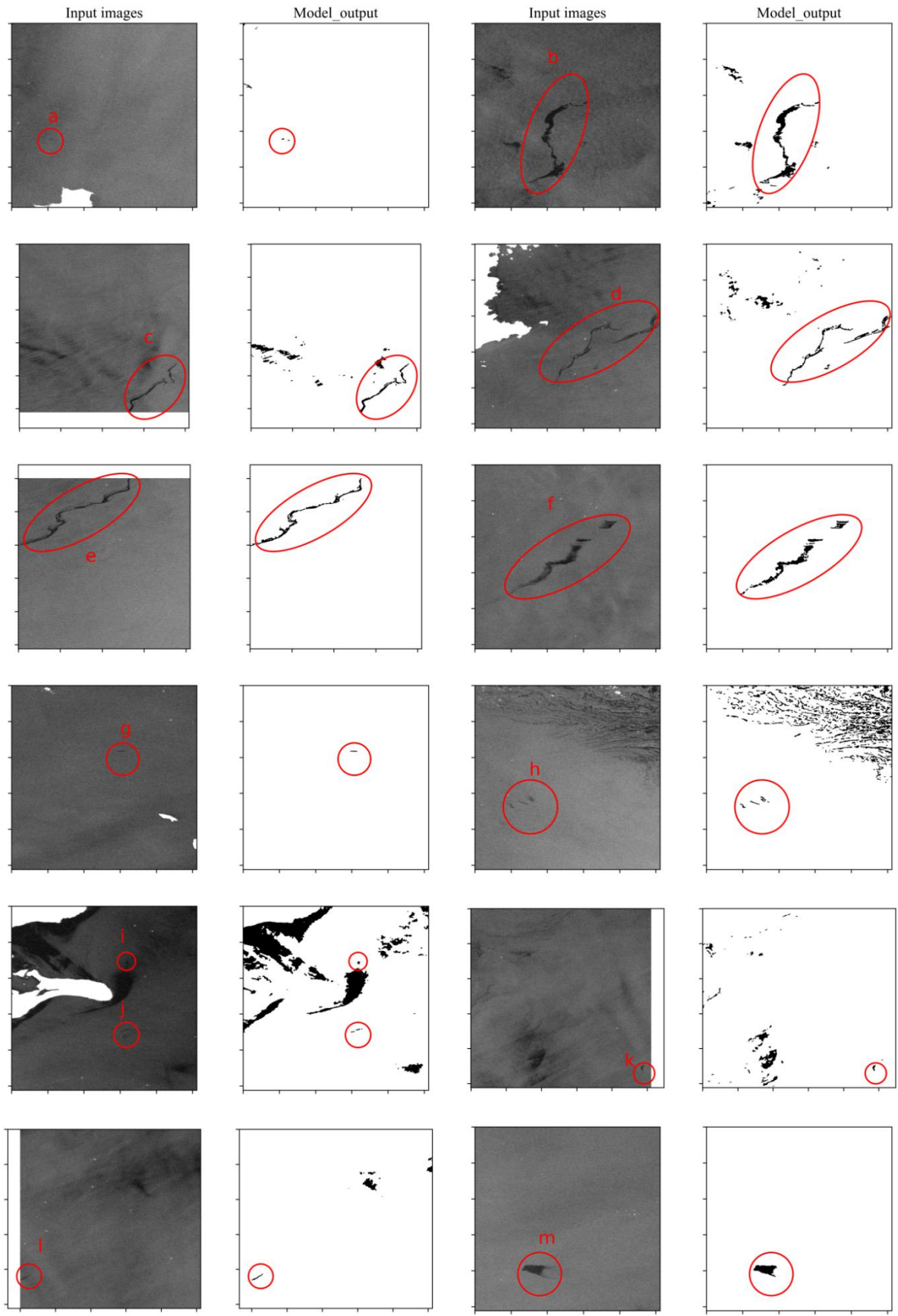

Figure 5. The results of successful segmentation of the oil patches of various shapes and sizes; the dark spots in the red circles are the oil patches and a ~ m are their numbers.

In addition to oil slicks, many meteorological or oceanic phenomena produce weak backscatter, which also appears as dark spots on SAR images (Shu et al. 2010); and these non-oil dark spots account for the vast majority of dark spots (Topouzelis 2008). Najoui et al. (2018) explained that the meteorological or oceanic phenomena that cause non-oil dark spots to appear on SAR images, include upwelling, rain cells, wind shadowing, and sea currents. Alpers et al. (2017) pointed out that areas with higher chlorophyll-a distribution in the ocean also show as dark spots on SAR images. Figure 6 shows several examples of various look-alikes successfully segmented. Table 6 shows the atmospheric and ocean surface characteristics of these look-alikes, including wind speed, ocean currents, chlorophyll-a concentration, and temperature difference between atmosphere and ocean. Except for the wind speed, which we derived using the CMOD5 model, other meteorological and marine data were provided by the European Center for Medium-Term Weather Forecast (ECMWF).

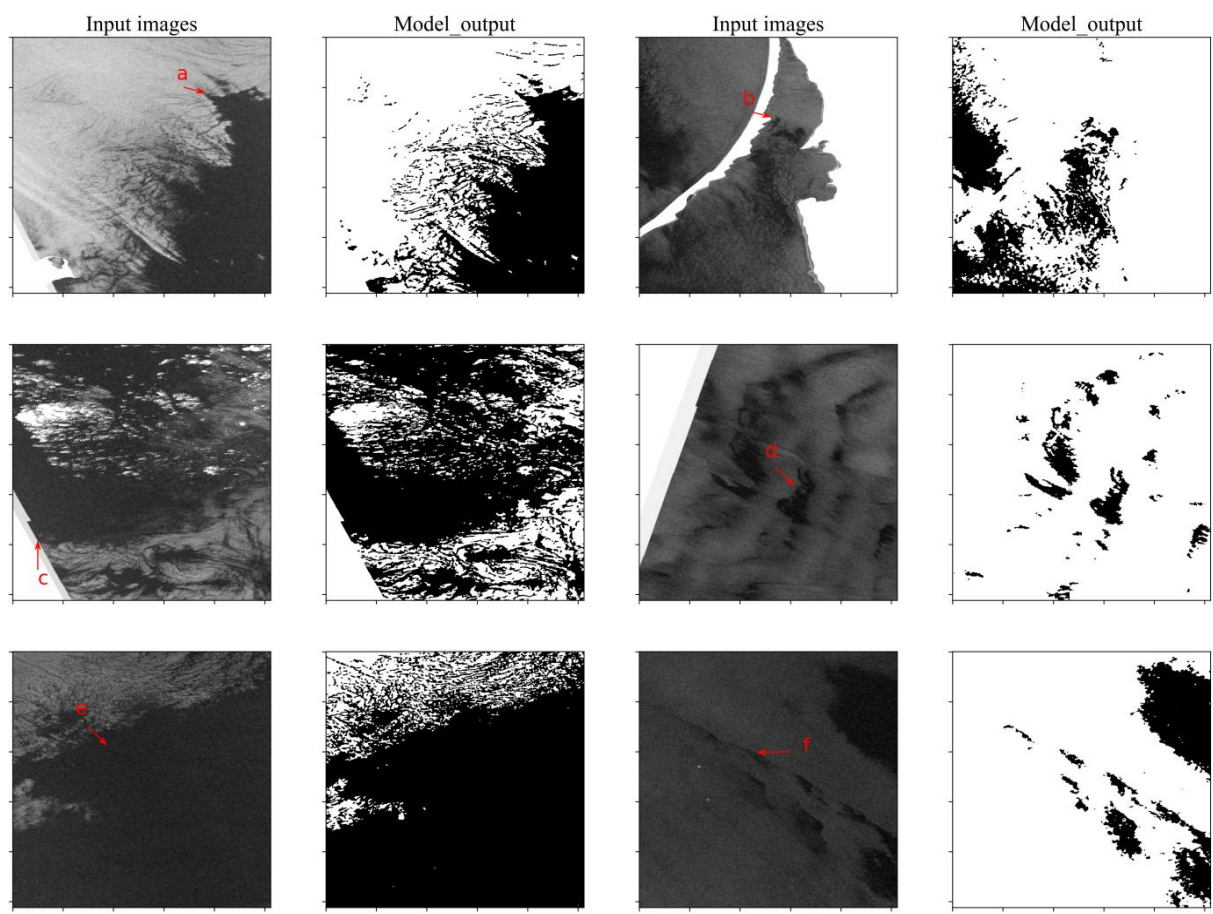

Figure 6. The segmentation results of various look-alikes; the a~f arrows point to the dark spots.

Table 6. Ocean and atmospheric characteristics of the look-alikes in Figure 8.

| Dark spot | Mean wind (m/s) | Mean sea water velocity (m/s) | Mean convective rain rate (kg*m$^{-2}$ *s$^{-1}$) | The temperature difference between the atmosphere and the ocean (K) | Mean Chlorophyll-a concentration (mg /m$^3$) |
|---|---|---|---|---|---|
| a | 0.195 | 0.088 | 0 | 0.635 | 2.128 |
| b | 4.009 | 0.040 | 0 | -0.704 | 17.400 |
| c | 0.289 | 0.058 | 0 | 0.201 | 2.459 |
| d | 4.750 | 0.083 | 0 | 3.655 | 0 |
| e | 0.553 | 0.078 | 0 | 0.144 | 2.635 |
| f | 4.012 | 0.206 | 0 | -0.689 | 2.673 |

In Figure 6, dark points a, c, and e are the low wind areas where the wind speed was lower than 0.6 m/s. The low wind speed resulted in a smooth sea surface so these areas appeared as dark spots on the images. Dark spots caused by low winds accounted for the vast majority of look-alikes on the images. The concentration of chlorophyll-a in the dark spot b area was 17.400 mg/m$^3$, which is relatively high. This dark spot may have been caused by an abnormal chlorophyll-a concentration. The temperature difference between the atmosphere and the ocean in the dark spot d area was relatively large at 3.655k. This dark spot may have been caused by an upwelling, which brings the cold water from the lower layer to the upper layer. On radar image, the temperature drop is usually followed by a decrease in the roughness of the sea surface, which appears as a dark spot (Najoui et al. 2018). The dark spot f may have been caused by the sea currents, which can cause biogenic oil to accumulate in some regions and change the roughness of the sea surface, making these areas appear as black spots on SAR images (Najoui et al. 2018). As far as the various causes and shapes of look-alikes, it can be seen that the segmentation results of the SDGCN model were basically the same as the dark spots on the input image. The segmentation results of this model were nearly satisfactory.

Although our method improves the performance of dark spots segmentation, it requires several complicated steps to implement, including transforming images to graphs with superpixels as nodes, feature extraction and selection of nodes, and graph node classification. Overall, it is time-consuming, especially in superpixel segmentation and feature extraction. In superpixel segmentation, there are also still a small number of dark spots with small areas and weak borders, whose contours are difficult to detect, resulting in missing detection. Furthermore, due to the small amount of training data, the model is over-fitted, which also leads to the lack of some dark spots in the detection. Therefore, it is necessary to develop superpixel segmentation algorithms with higher performance to improve the accuracy of contour detection. And more data needs to be used to train the model to reduce the effect of overfitting in practical applications. Dark spots detection is only the beginning of oil spill detection. In the future, we will utilize detected dark spots to build a knowledge graph to help identify oil slicks.

## 6 Conclusion

In this paper, we proposed an effective method of dark spot segmentation that can accurately detect dark spots on a single-polarized SAR image. Our method consists of three steps: 1) converting images to graphs with superpixels as nodes; 2) feature extraction and selection of superpixels; and 3) graph node classification. To perform superpixel segmentation of the image, our method uses the BASS algorithm, which can accurately detect the contours of the fuzzy dark spots and smooth the noise of the SAR image at the same time. Then, the superpixel is used as the basic unit and the image is transformed into a graph structure with its superpixels as nodes, which significantly reduces the computational burden. A feature vector composed of 137 feature values is calculated for each node, and an optimal feature subset consisting of 30 feature values is selected using the SVM-RFE algorithm for dark spot segmentation. The popular DeeperGCN classifier was chosen for graph node classification in our method. We found that the proposed feature subset not only accelerated the feature extraction process, but also improved the accuracy of the model. Among the selected features, the physical features played a major role, followed by the geometric features and the textural features. The experimental results show that our method is more effective than pixel-based segmentation methods and can segment most of the dark spots except for those with smaller shapes and brightness levels that were very close to the background brightness. Due to the general characteristics of SDGCN model, it can be easily extended to address applications such as semantic segmentation of optical remote sensing images (Li et al. 2020b). In the future work, we therefore intend to explore that possibility. And we will continue the follow-up work on dark spot detection. The dark spots segmented from the image will be used as entities to create a knowledge graph for oil spill detection (Hao et al. 2021). Then, we will explore knowledge inference methods to identify oil slicks.

## Acknowledgments

This work was supported by the National Key Research and Development Program of China under grant number 2018YFB0505003 and the National Natural Science Foundation of China under grant number 42030102. The authors are very grateful to Guohao Li, Roy Uziel and David Mera for code support and the ESA and the Baltic Marine Environment Protection Commission (Helsinki Commission - HELCOM) for data support.

## References

Alpers, W., Holt, B., & Zeng, K. (2017). Oil spill detection by imaging radars: Challenges and pitfalls. *Remote Sensing of Environment, 201*, 133-147

Berry, A., Dabrowski, T., & Lyons, K. (2012). The oil spill model OILTRANS and its application to the Celtic Sea. *Mar Pollut Bull, 64*, 2489-2501

Cantorna, D., Dafonte, C., Iglesias, A., & Arcay, B. (2019). Oil spill segmentation in SAR images using convolutional neural networks. A comparative analysis with clustering and logistic regression algorithms. *Applied Soft Computing, 84*

Chehresa, S., Amirkhani, A., Rezairad, G.-A., & Mosavi, M.R. (2016). Optimum Features Selection for oil Spill Detection in SAR Image. *Journal of the Indian Society of Remote Sensing, 44*, 775-787

ESA (2007). ASAR Product Handbook, 94-97

Feng, J., Chen, H., Bi, F., Li, J., & Wei, H. (2014). Detection of oil spills in a complex scene of SAR imagery. *Science China Technological Sciences, 57*, 2204-2209

Fingas, M., & Brown, C.E. (2017). A Review of Oil Spill Remote Sensing. *Sensors (Basel), 18*

Garcia-Pineda, O., Zimmer, B., Howard, M., Pichel, W., Li, X., & MacDonald, I.R. (2009). Using SAR images to delineate ocean oil slicks with a texture-classifying neural network algorithm (TCNNA). *Canadian Journal of Remote Sensing, 35*, 411-421

Genovez, P., Ebecken, N., Freitas, C., Bentz, C., & Freitas, R. (2017). Intelligent hybrid system for dark spot detection using SAR data. *Expert Systems with Applications, 81*, 384-397

Giraud, R., Ta, V.-T., & Papadakis, N. (2018). Robust superpixels using color and contour features along linear path. *Computer Vision and Image Understanding, 170*, 1-13

Guo, H., Wei, G., & An, J. (2018). Dark Spot Detection in SAR Images of Oil Spill Using Segnet. *Applied Sciences, 8*

Habart, D., Kybic, J., Švihlík, J., & Borovec, J. (2017). Supervised and unsupervised segmentation using superpixels, model estimation, and graph cut. *Journal of Electronic Imaging, 26*

Hao, X., Ji, Z., Li, X., Yin, L., Liu, L., Sun, M., Liu, Q., & Yang, R. (2021). Construction and Application of a Knowledge Graph. *Remote Sensing, 13*

Hersbach, H., Stoffelen, A., de Haan, S.J.E., amp, & Symposium, E. (2005). The Improved C-Band Geophysical Model Function CMOD5. In  (p. 142.141)

Kipf, T.N., & Welling, M. (2017). Semi-supervised classification with graph convolutional networks. *In International Conference on Learning Representations(ICLR)*

Konik, M., & Bradtke, K. (2016). Object-oriented approach to oil spill detection using ENVISAT ASAR images. *ISPRS Journal of Photogrammetry and Remote Sensing, 118*, 37-52

Lang, H., Zhang, X., Xi, Y., Zhang, X., & Li, W. (2017). Dark-spot segmentation for oil spill detection based on multifeature fusion classification in single-pol synthetic aperture radar imagery. *Journal of Applied Remote Sensing, 11*

Lee, J., Lee, I., & Kang, J.J.A. (2019). Self-Attention Graph Pooling*, abs/1904.08082*

Li, G., Muller, M., Thabet, A., & Ghanem, B. (2019). DeepGCNs: Can GCNs Go As Deep As CNNs? *2019 IEEE/CVF International Conference on Computer Vision (ICCV)*, 9266-9275

Li, G., Xiong, C., Thabet, A.K., & Ghanem, B.J.A. (2020a). DeeperGCN: All You Need to Train Deeper GCNs*, abs/2006.07739*

Li, Y., Chen, W., Zhang, Y., Tao, C., Xiao, R., & Tan, Y. (2020b). Accurate cloud detection in high-resolution remote sensing imagery by weakly supervised deep learning. *Remote Sensing of Environment, 250*

Li, Y., & Li, J. (2010). Oil spill detection from SAR intensity imagery using a marked point process. *Remote Sensing of Environment, 114*, 1590-1601

Liu, M., Gao, H., & Ji, S. (2020). Towards Deeper Graph Neural Networks. In, *Proceedings of the 26th ACM SIGKDD International Conference on Knowledge Discovery & Data Mining* (pp. 338-348)

Mera, D., Bolon-Canedo, V., Cotos, J.M., & Alonso-Betanzos, A. (2017). On the use of feature selection to improve the detection of sea oil spills in SAR images. *Computers & Geosciences, 100*, 166-178

Mera, D., Cotos, J.M., Varela-Pet, J., & Garcia-Pineda, O. (2012). Adaptive thresholding algorithm based on SAR images and wind data to segment oil spills along the northwest coast of the Iberian Peninsula. *Mar Pollut Bull, 64*, 2090-2096

Misra, A., & Balaji, R. (2017). Simple Approaches to Oil Spill Detection Using Sentinel Application Platform (SNAP)-Ocean Application Tools and Texture Analysis: A Comparative Study. *Journal of the Indian Society of Remote Sensing, 45*, 1065-1075

Najoui, Z., Riazanoff, S., Deffontaines, B., & Xavier, J.-P. (2018). A Statistical Approach to Preprocess and Enhance C-Band SAR Images in Order to Detect Automatically Marine Oil Slicks. *IEEE Transactions on Geoscience and Remote Sensing, 56*, 2554-2564

Rong, Y., Huang, W., Xu, T., & Huang, J. (2020). DropEdge: Towards Deep Graph Convolutional Networks on Node Classification. *In International Conference on Learning Representations (ICLR)*

Ronneberger, O., Fischer, P., & Brox, T. (2015). U-Net: Convolutional Networks for Biomedical Image Segmentation. *Medical Image Computing and Computer-Assisted Intervention – MICCAI 2015* (pp. 234-241)

Shu, Y., Li, J., Yousif, H., & Gomes, G. (2010). Dark-spot detection from SAR intensity imagery with spatial density thresholding for oil-spill monitoring. *Remote Sensing of Environment, 114*, 2026-2035

Solberg, A.H.S. (2012). Remote Sensing of Ocean Oil-Spill Pollution. *Proceedings of the IEEE, 100*, 2931-2945

Solberg, A.H.S., Brekke, C., & Husoy, P.O. (2007). Oil Spill Detection in Radarsat and Envisat SAR Images. *IEEE Transactions on Geoscience and Remote Sensing, 45*, 746-755

Solberg, A.S., Brekke, C., Solberg, R., & Husoy, P.O. (2004). Algorithms for oil spill detection in Radarsat and ENVISAT SAR images. *IGARSS 2004. 2004 IEEE International Geoscience and Remote Sensing Symposium*, 4909-4912

Taravat, A., Latini, D., & Del Frate, F. (2014). Fully Automatic Dark-Spot Detection From SAR Imagery With the Combination of Nonadaptive Weibull Multiplicative Model and Pulse-Coupled Neural Networks. *IEEE Transactions on Geoscience and Remote Sensing, 52*, 2427-2435

Topouzelis, K., Karathanassi, V., Pavlakis, P., & Rokos, D. (2006). Dark formation detection using recurrent neural networks and SAR data. In, *Image and Signal Processing for Remote Sensing XII*

Topouzelis, K., Singha, S., & Kitsiou, D. (2016). Incidence angle normalization of Wide Swath SAR data for oceanographic applications. *Open Geosciences, 8*, 450-464

Topouzelis, K.N. (2008). Oil Spill Detection by SAR Images: Dark Formation Detection, Feature Extraction and Classification Algorithms. *Sensors (Basel), 8*, 6642-6659

Uziel, R., Ronen, M., & Freifeld, O. (2019). Bayesian adaptive superpixel segmen. *2019 IEEE/CVF International Conference on Computer Vision (ICCV)*, 2019, pp. 8469-8478

Vyas, K., Shah, P., Patel, U., Zaveri, T., & Kumar, R. (2015). Oil spill detection from SAR image data for remote monitoring of marine pollution using light weight imageJ implementation. *2015 5th Nirma University International Conference on Engineering (NUiCONE)*, 1-6

Xu, L., Shafiee, M.J., Wong, A., & Clausi, D.A. (2016). Fully Connected Continuous Conditional Random Field With Stochastic Cliques for Dark-Spot Detection In SAR Imagery. *IEEE Journal of Selected Topics in Applied Earth Observations and Remote Sensing, 9*, 2882-2890

Yu, X., Zhang, H., Luo, C., Qi, H., & Ren, P. (2018). Oil Spill Segmentation via Adversarial $f$ -Divergence Learning. *IEEE Transactions on Geoscience and Remote Sensing, 56*, 4973-4988

Zhang, M., & Chen, Y. (2018). Link prediction based on graph neural networks. In, *Proceedings of the 32nd International Conference on Neural Information Processing Systems* (pp. 5171–5181). Montréal, Canada: Curran Associates Inc.